# Constant for associative patterns ensemble

Leonid Makarov and Peter Komarov[1]
*St. Petersburg State University of Telecommunications,
Bolshevikov 22, 193382, St. Petersburg, Russia*

**ABSTRACT**

Creation procedure of associative patterns ensemble in terms of formal logic with using neural network (NN) model is formulated. It is shown that the associative patterns set is created by means of unique procedure of NN work which having individual parameters of entrance stimulus transformation. It is ascertained that the quantity of the selected associative patterns possesses is a constant.
**Keywords:** *computer science, neural network (NN), associative pattern.*

## 1. INTRODUCTION

The set of the natural-science facts which validity cannot be established is known. Such representation about various events and the phenomena which are found out in world around constantly is exposed to check. Using the knowledge received in different areas of research, theoretical models which approach to understanding of process essence or the phenomenon are created. Approximation of model researches results on the certain class of the phenomena allows constructing the axioms which accepted at once and without proofs. The concept "axiom" is founded on logic irreducibility. Axioms are mathematical positions which rely self-evident and do not demand proofs. At the axioms formulation rather simple and clear principles making results of numerous supervision or researches, executed are used by different methods. It is recognized, that mathematical theories are based on axioms from which consequences which name theorems are deduced.

The living organism's ability of accumulating the information about events is the axiomatic fact. Principles of the memory block working of a computer the Von Neumann architecture and a human brain has the certain similarities and distinctions. Representations about an organism NN allows to believe that information blocks of messages are constructed of elements - elementary information acts of complex system (a brain) conditions change. This common sight at complex working processes of the computer's and the human's memory block with information streams allows to approve that the combination of elements creates set of the information blocks differing on quantitative and qualitative measures. Among created information blocks are available such which have the certain similarity – they possessed associativity.

Computer uses the address in data file for search of the set information, and the human uses association. Therefore, if the information address is known computer program will find the established information block. In that case when the address is unknown, the search computer program needs to carry out full viewing of all information blocks. In this procedure the case of full concurrence of the established and required information blocks is by default examined only. Even small variability in information blocks comparison creates increased requirements to the search computer program. Search of the human – is reproduction of the demanded information block in memory and it occurs on the basis of the address to associative memory. Organization of this memory is based on representation that all elements of the information block are connected among themselves is associative by a position principle of an arrangement of elements in the information block. Research of associative memory can be spent on the Hamming network model having a set of input and output parameters. Input parameters set of model are designate a stimulus vector:

$$X\{x_1, x_2, ..., x_i, ..., x_n\}, \ 0 \leq x \leq 1 \qquad (1)$$

and output parameters set is observation vector:

$$Y\{y_1, y_2, ..., y_i, ..., y_n\}, \ 0 \leq y \leq 1 \qquad (2)$$

---

The algorithm of network model work is based on associative patterns affinity definition. It is identical to definition of an entrance and target vectors conformity $X \cong Y$. The affinity of associative patterns can be estimated by Hamming metric which determines distance as differing positions quantity of elements in vectors. Result of network work is the finding of pattern with the least distance in relation to the base pattern accepted as a basis.

Using the common concepts of real events perception in Hamming model the associativity is considered for two group of events $X$ and $Y$. Considering $Y$ as a pattern received as a result of special rules transformation of an entrance pattern $X$ application it is possible to create a series of associative patterns:

$$Q_j, 1 \leq j \leq m \qquad (3)$$

In this case a series of associative patterns is created in view of neural network specific features.

Process of purposeful search of close vectors can be considered as extraction and the analysis of information reports in the communications. This problem well-known for computer science problems which were formed on the basis of many scientific disciplines considering problems of thinking, constructions of logic automatic devices for a judgments conclusion, formations of the knowledge bases. Constructing of reasonable-cognitive mediums reflecting powers of human thinking, is directly connected with modeling of complex neural structures work. Accordingly the artificial intelligence (AI) is usually treated how property of computing systems to execute separate functions of human intelligence, for example, to choose and make optimum decisions on the basis of earlier received experience - of the knowledge base and the rational analysis of external action. The system allocated by intelligence is a universal means of the wide range problems decision, including not formalized for which there are no standard, in advance known methods of the decision.

## 2. NEURONAL INFORMATION BASIS OF AN ASSOCIATIVE PATTERN

Modern tendencies of active development and all-round using computer engineering means, in various areas of human practical activities, have prevailing character. On the one hand it is directly bounded up with growth of solved problems complexity in the research and practice field, on the other hand – with natural aspiration to reproduce more comfortable mediums of human and infomedia interaction. Hardware using including such which provides conversational procedures carrying out has considerably changed of the communication act character. Human representation about outward things becomes more and more mediated act which organized and supported by hardware-software means complex aggregate. The modern interaction interface "human – computer" is constructed on the basis of information and biophysical concepts about the communications. The computing environment created by a computer, is considered as means of the communicative patterns creation which based on visual, acoustic and olfactory human ability to perceive objects of the real world. Processing technique using in the communication acts are formed the specific environment where reproduced patterns are exposed to definitely expressed modification.

The reasoning logic scheme of reception of an associative pattern is build on an axiom, that relative, to formal attributes stimulus vectors $X_i \cong X_{i+1}$ create in NN model almost adequate observation vectors $Y_i \cong Y_{i+1}$. The Hamming NN model is similar to complex system, for example, to a living organism NN. Then pattern associativity is established on the ground that set of input action on the system, having established parameters value, is a subset of complex system output responses.

The mainstream of approaches in neuro-information science is formed from a position of neural schemes methods creation. These schemes are solving those or other tasks, in particular, such as patterns which arise after stimulus reception. The perception of visual, acoustic and aromatic stimulus generates a pattern which, being reproduced on the communication network terminal creates preconditions for the communication act development. Representation of information package pattern, which reproduced on the communication network terminal, is starting reception process realized by NN and concluded by pattern formation. In such conception the communication pattern occurrence is directly linked with sensor NN work. The neuron model using is appears useful by general principles consideration of the information package pattern formation.

Distinctions between approaches and the separate neuron and a NN work studying methods, generally, can be recognized not so essential when the general principles of the data analysis and creation of a modeling pattern of information package perception are considered. In such understanding of the device neuro-information science are identical to communication system where the greatest interest is represented with problems in formation of information packages of messages in habitats patterns perception - an organism of the human and environments artificial - computing.

Using natural preconditions for an represented processes at a level of modules of logic construction of the conclusions organized on NN structure and processes proceeding at a level of organism touch systems isomorphism establishment, we shall put, that the message information package pattern, it is presented by a set of elements $\{x_1, x_2, x_3, ..., x_n\}$.

Believing that $X \Rightarrow Y$ it is possible to spend comparisons of input factors set to output factors which are identified with associative pattern $Q$ creation procedure. Formally, supposing existence of several associative patterns, it is possible to write down: $Y \Rightarrow Q_j$. It appears from this that the set of in parameters $X$ creates a set of associative patterns $Q_j$ which are constructed by means of elements $Y$ positions combination. Is marked the set of patterns $Q_j$ contains the common elements located on certain positions. The sets partially containing the common elements refer to overlapping sets. Such sets we shall determine as associative patterns. Using representations about formation of an events pattern $Q$, we shall establish the associative patterns formation general principles.

Let's put, that some certain element of set $Q_2$ is put to each element of set $Q_1$ in conformity by virtue of, any rule or the law. If thus each element of set appears put in conformity to one and only to one element of other set between sets $Q_1$ and $Q_2$ it is established one-to-one correspondence. It is significant that two elements participate in procedure of conformity establishment between sets, on one of each of the presented sets at least. It is obvious that the given concept of sets comparison can be expanded and considered more than two elements forming definitely allocated groups - clusters. In that case comparison also will pass in pairs, but this procedure will be already executed at elements groups' level. We shall note, that is formal, the elements quantity which entering into allocated groups at no allowance. Procedure of sets comparison allows establishing both full and partial sets conformity. This procedure under the general execution scheme is close to of Hamming metrics calculation procedure.

### 3. ASSOCIATIVE PATTERN CONSTRUCTION METHOD

We shall distribute to NN the given reasoning. Let there are two sets $X$ and $Y$. We believe that some certain element $y = f(x)$ of set $Y$ is put to each element $x \in X$ in conformity. In that case we can assert that there is mapping of set $X$ in set $Y$. In other words there is a function which argument $x$ is realized on all on set $X$, and values $y$ belong to set $Y$. It appears from this that for each datum $x \in X$ the element $y = f(x)$ of set $Y$ is pattern of an element $x \in X$. In this case the associative patterns set are defined in the form of:

$$|X \cup Y| = |X| + |Y| - |X - Y| \qquad (4)$$

In the general case when a series $m$ from sets is considered, this expression looks in the following way:

$$|A_1 \cup A_2 \cup ... \cup A_m| = \sum_i |A_i| - \sum_{i<j} |A_i A_j| + \sum_{i<j<k} |A_i A_j A_k| - ... + (-1)^{m+1} (|A_1 A_2 ... A_m|) \qquad (5)$$

where indexes $i, j, k$ change from $1$ up to $m$. It is obvious that at $m = 2$ we receive expression (4). The analysis of expression (5) allows establishing quantity composed $\Psi$ if the quantity $m$ of initial sets is known:

$$\Psi = 2^m - 1 \qquad (6)$$

It is obvious that with growth $m$ the quantity items $\Psi$ in expression (5) increases. So, for example, at $m = 40$ the quantity of items members $\Psi$ in expression (5) will exceed billion. It is necessary to recognize, that such volume of calculations is difficult for executing in practice. Taking it into consideration, we shall take advantage of an induction method and we shall believe that expression (5) is evenly for a case $m' = 2$ and $m' = m - 1$.

Let's put

$$A_1 \cup ... \cup A_{m-1} = B \qquad (7)$$

Then expression (5) will become:

$$|A_1 \cup ... \cup A_m| = |B \cup A_m| = |B| + |A_m| - |BA_m| = |B| + |A_m| - |A_1 A_m \cup ... \cup A_{m-1} A_m| =$$
$$= \sum_{i<m} |A_i| + |A_m| - \sum_{i<j<m} |A_i A_j| - \sum_{i<m} |A_i A_m| + ... = ... \sum_i |A_i| - \sum_{i<j} |A_i A_j| + ... \qquad (8)$$

The received result allocates the pattern elements sets and corresponds to expression (5). The result we shall demonstrate with an example. We shall put, there is a set:

$$X \Rightarrow Y \Rightarrow Q_i = e\{e_1, e_2, e_3\} \qquad (9)$$

Change of the elements following order in the specified set creates a series from $m$ associative patterns $m = n! = 6$:

$$\begin{aligned} e^1 &= \{a_1, a_2, a_3\} \\ e^2 &= \{a_1, a_3, a_2\} \\ e^3 &= \{a_2, a_3, a_1\} \\ e^4 &= \{a_2, a_1, a_3\} \\ e^5 &= \{a_3, a_1, a_2\} \\ e^6 &= \{a_3, a_2, a_1\} \end{aligned} \qquad (10)$$

In the specified pattern series, four combinations $(\chi = 4)$ where elements keep a starting position are easily found out. For the selected elements set operation of pair comparison $l = h$ is spent. In a set $e^1$ two combinations are included, in a set $e^3$ one combination is included and in a set $e^5$ one combination is included. It is obvious that at increase in dimension of parameters set it is required to spend operation of comparison $l \leq h \leq (n-1)$. According to the established principles of an assortment from 6 associative patterns creation it is had:

$$e \Rightarrow Q(e^1, e^2, e^3, e^4, e^5, e^6) \qquad (11)$$

Thus, a necessary condition of associative patterns set creation is a vector $Y$ presence. Dimension of a vector $Y$ defines the associative patterns quantity.

**Let's prove a lemma.**

The associative patterns set created on the basis of a vector $Y$ contain $\chi$ groups of motionless elements with natural sequence. The quantity of groups with natural elements sequence is boundedly and does not surpass a constant $\mu$.

**Lemma proving.**

Using the entered concept, we shall examine a task with finding of the sets quantity made of motionless elements in an associative pattern. For the examined case at $n = 3$ it is had $\chi = 4$. For $n = 2$, with all evidence it is received: $\chi = 1$.

Let's note that using duplicating procedure of an initial combination $e = \{a_1, a_2, ..., a_n\}$ by change of the elements sequence new event patterns is created.

Thus by virtue of some formal rules or the characteristic attributes uniting given set elements, the created pattern series appears the interconnected train - having the common logic term. Creation of a pattern series can be determined as rearrangement procedure of the initial pattern elements, generating pattern creation which having motionless elements.

Let's examine the general reasoning scheme of compositions detection procedure search in which elements change a starting position in a train. We shall put there is a set of rearrangements $A_k$ with a motionless element $k$, such that $1 \leq k \leq m$. It is required to find pattern quantity with motionless elements: $N_m = |A_1 \cup ... \cup A_m|$. Let's note that for any $k$ equality is carried out:

$$\left| A_{i_1}, A_{i_2}, ..., A_{i_k} \right| = (m-k)! \tag{12}$$

Really the set crossings organization it is possible to examine as the initial set modification made of the ordered set of elements. Then the set $A_{i_1} ... A_{i_k}$ is identified with a set of rearrangements multitudes in which elements with numbers $i_1, ..., i_k$ are fixed on the initial positions, and the others $m-k$ of elements from $1, ..., n$ positions can be located in the any order on $m-k$ free positions. It is formally possible to not bring to a focus to $m-k$ elements and their hit on initial positions. In that case expression (5) can be simplified by means composed quantity change in the sum expression where there are crossings a kind $A_{i_1} ... A_{i_k}$. Then:

$$\binom{m}{k} = \frac{m!}{k!(m-k)!} \tag{13}$$

Considering this, composed in expression (5) it is possible to replace everyone with expression $(m-k)!$. We receive that the combinations with motionless elements quantity is determined in the form of:

$$N_m = |A_1 \cup A_2 \cup ... \cup A_m| = \frac{m!}{1!} - \frac{m!}{2!} + \frac{m!}{3!} - ... + (-1)^{m+1} \tag{14}$$

The received expression (14) establishes the associative patterns quantity possessing motionless elements. Thus it is possible to note that the established estimation $N_m$ depends on the size of an initial element set $\{y_1, y_2, ..., y_i, ..., y_n\}$ making an initial pattern. The duplicating of an initial pattern, examined as procedure of associative pattern train construction, some of which possess property of separate elements selective positioning, allows examining a ratio:

$$\Psi_m = \frac{N_m}{m!} = \frac{1}{1!} - \frac{1}{2!} + \frac{1}{3!} - ... + \frac{(-1)^{m+1}}{m!} \tag{15}$$

The analysis of a ratio (15) appears useful and allows establishing that with growth $m$ the estimation $\Psi_m$ comes nearer to the infinite sum $S_m$:

$$S_m = \frac{1}{1!} - \frac{1}{2!} + \frac{1}{3!} - \frac{1}{4!} + ... = \pm \frac{1}{m_i!} \tag{16}$$

where: $1 < i < n!$. The given sum series converges:

$$\lim_{m \to \infty} S_m = 1 - e^{-1} = 0.6321 \tag{17}$$

**Lemma is proved.**

The received expression establishes that with increase $m$ the associative patterns $N_m$ quantity, which possessing motionless elements, increases, but their relative share does not surpass value $\mu = 0.6321$.

## 4. RESULTS AND DISCUSSION

The received value $\mu$ establishes that in not to dependence on volume (the size) of an initial pattern for any kind of messages package the share of patterns with motionless elements remains to a constant. From a position of the computer calculations organization the control only this pattern train part allows to lower volume of computing procedures. Expanding these representations it is possible to specify that artificial distortion or reduction of a train set $\chi$ will inevitably lead to easing of initial messages (the patterns) recognition. The given representation in the communication theory is well-known. Using of the entered concepts about associative patterns formation is well-formed with representations about neural structures long-term memory. This remark can is useful at complex systems and databases creation.

Development of modern communications means is indissolubly connected with perfection of approaches in constructing the interfaces providing a high overall performance of computing environments, possessing elements of the artificial intellect, natural abilities of the human promoting rapprochement to form and recognize complex communication patterns. In this process win first place the associative patterns which are required to be reproduced in the computing environment on the modeling representations basis generated in view of neuro-information science. The prediction model construction, with use of combinatory mathematics, allows revealing the most essential parameters of associative patterns, to specify the general approaches to databases formation rules construction. It is quite natural that the given remark allows to lower requirements to volume of calculations in various recognition of the complex patterns tasks formed in view of touch human opportunities, capable to perceive on the communication terminal visual, acoustic or olfactory patterns.